\pdfoutput=1

\documentclass[11pt]{article}

\usepackage{acl}

\usepackage{times}
\usepackage{latexsym}

\usepackage[T1]{fontenc}

\usepackage[utf8]{inputenc}

\usepackage{microtype}
\usepackage{float}

\usepackage{color}
\usepackage{hyperref}
\usepackage{amsmath, geometry, amssymb, fullpage, graphicx, amsthm}
\usepackage{enumitem}
\usepackage{caption}
\usepackage{subcaption}
\usepackage[rightcaption]{sidecap}
\usepackage{bm}
\usepackage{bbm}
\usepackage{titlesec}
\usepackage{adjustbox}
\usepackage{booktabs}
\usepackage[colorinlistoftodos]{todonotes}
\usepackage{wrapfig}

\title{Lacuna Reconstruction: Self-supervised Pre-training \\ for Low-Resource Historical Document Transcription}

\author{Nikolai Vogler \\
  Computer Science and Engineering \\
  University of California, San Diego \\
  \texttt{nvogler@ucsd.edu} \\\And
  Jonathan Parkes Allen \\
  Roshan Institute for Persian Studies \\
  University of Maryland \\
  \texttt{jallen22@umd.edu} \\\AND
  Matthew Thomas Miller \\
  Roshan Institute for Persian Studies \\
  University of Maryland \\
  \texttt{mtmiller@umd.edu} \\\And
  Taylor Berg-Kirkpatrick \\
  Computer Science and Engineering \\
  University of California, San Diego \\
  \texttt{tberg@ucsd.edu} \\}

\begin{document}
\maketitle
\begin{abstract}
We present a self-supervised pre-training approach for learning rich visual language representations for both handwritten and printed historical document transcription.
After supervised fine-tuning of our pre-trained encoder representations for low-resource document transcription on two languages, (1) a heterogeneous set of handwritten Islamicate manuscript images and (2) early modern English printed documents, we show a meaningful improvement in recognition accuracy over the same supervised model trained from scratch with as few as 30 line image transcriptions for training.
Our masked language model-style pre-training strategy, where the model is trained to be able to identify the true masked visual representation from distractors sampled from \textit{within the same line}, encourages learning robust contextualized language representations invariant to scribal writing style and printing noise present across documents.
\end{abstract}

\section{Introduction}\label{sec:intro}

Document transcription is the task of converting images of handwritten or printed text into a symbolic form suitable for indexing, searching, and computational analysis.\footnote{We use the generic term \textit{document transcription} to refer to both the task of optical character recognition (OCR), which is typically reserved for \textit{printed} documents, and handwritten text recognition (HTR) for manuscripts.}
Historical documents, whether they were (re)produced via handwriting or the early printing press, confound current statistical document transcription models due to (1) extremely varied style and content across domains, (2) the presence of noise, and (3) a dearth of labeled data.

\begin{figure}[t]
    \centering
    \frame{\includegraphics[trim=10 548 20 10,clip,width=\linewidth]{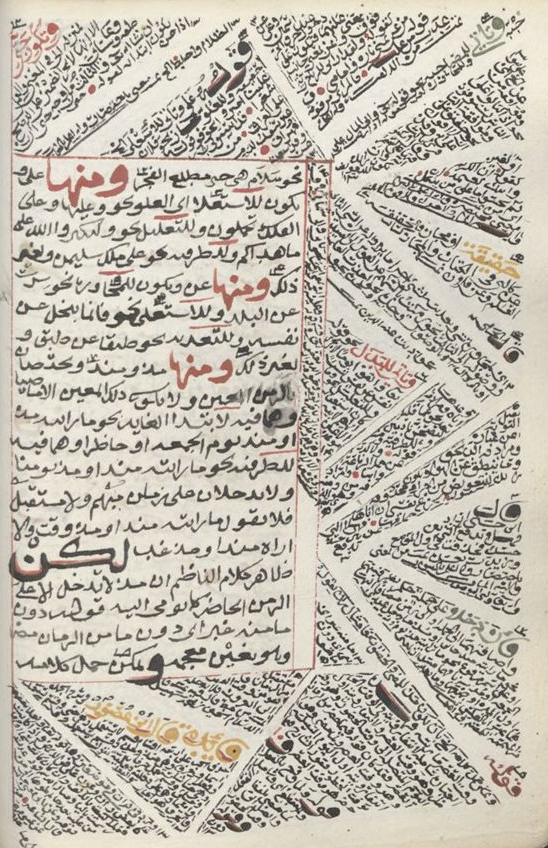}}
    \frame{\includegraphics[trim=10 548 0 5,clip,width=\linewidth]{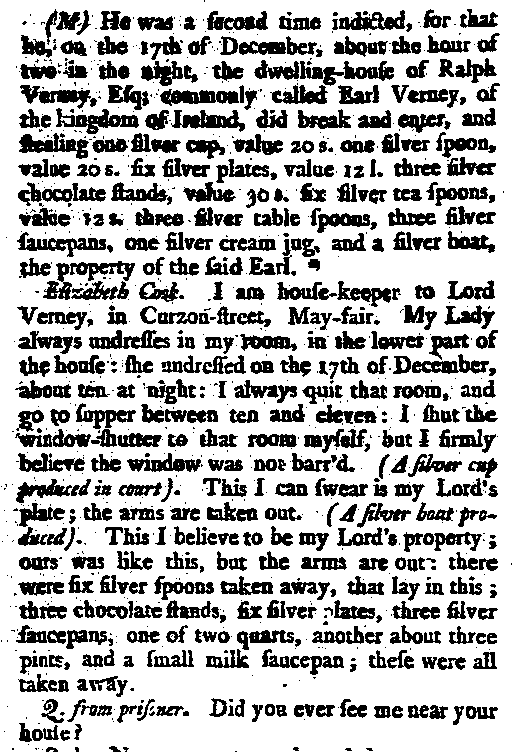}}
    \caption{Example page image crops from an Islamicate manuscript dated to 1842 (\textbf{Top}, ref: Leiden Or. 669), showcasing its dense, visual complexity with extensive marginalia, and printed proceedings of London's Old Bailey Courthouse (\textbf{Bottom}, c. 18\textsuperscript{th} century) \citep{shoemaker2005digital}.}
    \label{fig:example_pages}
\end{figure}

First, historical printed documents, such as books produced from early modern England (c. 16\textsuperscript{th}--18\textsuperscript{th} centuries; bottom of Fig.~\ref{fig:example_pages}), use non-standardized spacing and fonts \citep{shoemaker2005digital} and can contain code-switching that confuses language models \citep{garrette2015unsupervised}.
However, this variation pales in comparison to their handwritten counterparts.
For instance, pre-modern Islamicate manuscripts (i.e., Persian and Arabic handwritten documents from c. 7\textsuperscript{th}--19\textsuperscript{th} centuries; top of Fig.~\ref{fig:example_pages}), differ in script family, scribal handwriting style, and symbol inventory/vocabulary.
As a result, a large degradation in performance is observed when evaluating HTR models on unseen manuscripts \citep{jaramillo2018boosting}.

Production and imaging noise also present a problem for historical document transcription models.
Whether it be uneven inking from a printing press, inconsistent text baselines, or holes resulting from insect damage to ancient pages, techniques must be designed to cope with the noise \citep{berg2014improved,goyal2019global}.

While neural networks have a demonstrated capability to model complex data distributions, they typically require large amounts of supervised training data to do so, which is infeasible for historical documents.
Unsupervised, non-neural transcription models with fewer parameters alleviate the need to create labeled data \citep{berg2013unsupervised}, but struggle with complex handwriting variation.
For Islamicate manuscripts, ground truth transcription often requires paleography experts to decipher the ancient writing systems as they appear in each scribal writing style.

In this paper, we propose a self-supervised learning framework designed to overcome these three challenges presented by historical documents.
Inspired by the astounding success of self-supervised pre-training techniques for masked language modeling (MLM) in NLP \citep{devlin2019bert}, 
visual models \citep{chen2020simple,radford2021learning}, %
and speech recognition \citep{baevski2020wav2vec}, our approach pre-trains a neural text line-image encoder by learning to distinguish masked regions of unlabeled line images from other distractor regions. 
Specifically, our contribution is the following:
\begin{itemize}
    \item we show that the recent pre-train/fine-tune paradigm is particularly advantageous for low-resource historical document transcription, obtaining large improvements in both printed and handwritten documents in both English and Arabic-script languages.
    \item we motivate the self-supervised contrastive loss for document transcription through the lens of ``lacuna reconstruction'', where blank parts of a document called lacuna must be inferred by human readers.
\end{itemize}
In doing so, we argue that our approach to pre-training implicitly incentivizes the model to discover and encode discrete character classes in its internal representations, while ignoring style differences occurring in lines using different fonts,  languages, or authored by other scribes.

\section{Related Work}

\paragraph{Masked Pre-training}

Our approach to self-supervised pre-training follows a growing body of work in both NLP and speech that leverages mask-predict objectives for learning useful, task-agnostic language representations from unlabeled data.
In the self-supervised pre-train/supervised fine-tune paradigm, these representations can then be updated on the task of interest using in-domain labeled data.
Past work covers learning representations for NLP from monolingual and multilingual text \citep{devlin2019bert,yang2019xlnet}, speech \citep{baevski2019effectiveness,jiang2019improving,song2020speech,wang2020unsupervised}, and images grounded with text \citep{radford2021learning}.
Representations can be learned either through reconstruction objectives \citep{jiang2019improving,song2020speech,wang2020unsupervised} as opposed to a probabilistic contrastive loss \citep{oord2018representation,baevski2019effectiveness,baevski2020wav2vec}.
Most similar to our work is wav2vec2.0 \citep{baevski2020wav2vec}, which uses the same two phase training setup with a self-supervised contrastive loss during pre-training and Connectionist Temporal Classification (CTC) loss on transcribed speech data during fine-tuning.
\citet{talnikar2020joint} presents that the self-supervised loss regularizes the supervised loss during joint learning of both objectives.
Follow up work has shown the usefulness of the pre-trained speech representations for exploring speech variation \citep{bartelds2020neural}. 
In this paper, we show that the same learning paradigm can be also be successfully applied to much lower resource document transcription settings.

\paragraph{Islamicate HTR}
While machine recognition of handwritten, historic English/German documents can range from 5--12\% character error rate (CER) on a sufficient amount of in-sample manuscript training data \citep{sanchez_benchmarkhtr_2019}, performance on Arabic-script languages is much more challenging, leading to substantially higher CER.
Pre-modern Islamicate manuscripts (i.e., Persian and Arabic handwritten documents from c. 7\textsuperscript{th}--19\textsuperscript{th} centuries), often differ in script family, scribal handwriting style, and symbol inventory/vocabulary.
In the top of Figure~\ref{fig:example_pages}, we present an extreme example of some of the problematic visual variation that can be observed.
Even a model trained in a supervised fashion on such a complex document sees a large degradation in performance when evaluating HTR models on unseen manuscripts \citep{jaramillo2018boosting} .
Until quite recently, OCR performance on Arabic-script \emph{printed} texts was still quite poor, typically above 25\% CER \citep{alghamdi_experimental_2017}, which is still too high for downstream users (i.e., researchers and librarians). %

Recent studies involving Islamicate manuscripts found that state-of-the-art systems are only able to achieve 40 to mid-20\% CER using proprietary software (e.g., Google Cloud Vision, RDI, Transkribus) \citep{clausner_rasm_2018, keinanschoonbaert_rasm_2019, keinanschoonbaert_transkribus_2020}.
However, results from these studies only report in-domain performance---an unrealistic scenario where considerable amounts of labeled data can be obtained to enable both training and testing on the same manuscript.
In contrast, out-of-domain performance tends to suffer considerably, supported by \citet{romanov2017important}'s study of neural OCR for printed Arabic-script documents. 
Our work aims to address such performance issues for both in-domain and out-of-domain Islamicate HTR settings by learning general, content-rich pre-trained language representations from large amounts of heterogeneous unlabeled data.

\begin{figure*}[t]
    \centering
    \includegraphics[width=0.8\textwidth]{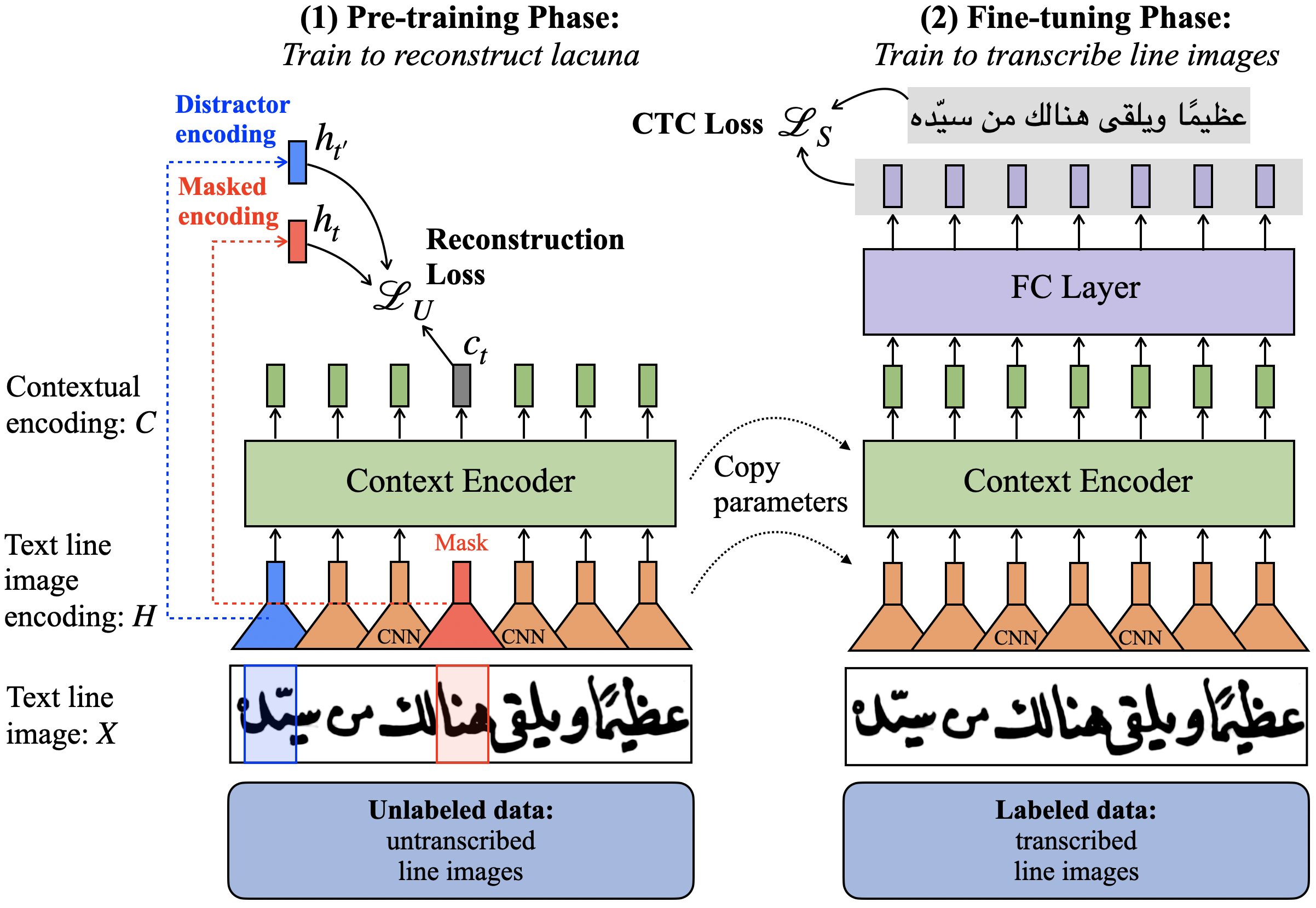}
    \caption{\small Our proposed two-stage approach for low-resource document transcription first pre-trains a line image encoder using a self-supervised contrastive loss on unlabeled data (left), followed by a fine-tuning phase, in which the pre-trained encoder learns to transcribe 1--3 pages of supervised data using a CTC loss (right).}
     \label{fig:pretrain_model_fig_basic}
\end{figure*}

\paragraph{Historical OCR}

Closely related to manuscript transcription, OCR is another task involving language recognition from images. 
However, OCR operates on documents that have been printed by a machine with regular, re-used character fonts exhibiting much less superficial glyph variation than human handwriting.
OCR is far from a solved problem in the case of documents printed on early modern (c. 16\textsuperscript{th}--18\textsuperscript{th} centuries; see bottom of Fig.~\ref{fig:example_pages}), movable-type printing presses, where humans would manually set metal type casts with non-standard spacing and fonts \citep{shoemaker2005digital}.
In this setting, inking noise and historical font shapes confuse OCR models trained on modern, computer-generated documents \citep{arlitsch2004microfilm}.
\citet{berg2013unsupervised}'s Ocular explicitly uses a generative probabilistic model inspired by historical printing processes to model such noise. 
Later work has extended it to handle more typesetting noise \citep{berg2014improved}, code-switched documents \citep{garrett_how_2014}, and produce both diplomatic and normalized transcriptions \citep{garrette2016unsupervised}.
Separately, OCR post-correction models have been proposed to resolve OCR outputs in historical documents \citep{hamalainen2019paft,dong2018multi} and other low-resource settings \citep{rijhwani2020ocr,rijhwani2021lexically}.
In contrast, our approach pre-trains the visual language recognition model's encoder, which  produces better contextualized representations in order to reduce the amount of errors the model itself makes.
Unlike Ocular, our proposed method does not use a language model and is not fully unsupervised as we require 1-3 pages of transcribed data for learning to transcribe during fine-tuning.

\section{Approach}

When human readers encounter a lacuna, a blank information gap in a portion of a book or manuscript, they must infer its latent meaning using nearby context like in a cloze test \citep{taylor1953cloze}.
We argue that the most useful information for inference lies in the ability to reason about the identities of the missing characters in the lacuna using the identities of the surrounding characters.
Indeed, MLM-style pre-training techniques are also motivated by the idea of the cloze test, and recent research indicates that language representations learned through the prediction of missing content using surrounding sentential context are useful for many downstream tasks \citep{devlin2019bert,clark2019electra,clark2020pre}.
Our approach combines the ideas of lacuna inference and masked pre-training to provide a useful learning signal for downstream historical document transcription, a setting with massive digitized collections but few transcribed examples.

Specifically, we introduce a self-supervised pre-training method that randomly masks lacuna-like regions of document line images and learns to reconstruct them by  distinguishing them from nearby line image segments, or foils.
While lacuna can be reconstructed in a generative way, we find that a discriminative contrastive loss works better in practice.
By leveraging a diverse set of unlabeled data for pre-training, the model is forced to infer the identities of masked text regions in the presence of scribal writing variation or typesetting noise ubiquitous in historical documents.
In the next sections we describe our model and masking strategy in more detail.

\subsection{Model}

In Figure~\ref{fig:pretrain_model_fig_basic}, we show our two-stage pre-train/fine-tune modeling approach.
First, we describe the document line image encoder that is shared between stages.
For simplicity of description, we assume that each document line image, $X$, is $n$ pixels tall and $m$ pixels wide, and that pixels are binary-valued. 
Thus, the space of input text line images can be denoted as $\mathcal{X} = \{0,1\}^{n \times m}$.
We first process the input with a \textbf{convolutional feature extractor}, $f: \mathcal{X} \mapsto \mathcal{H}$, that maps the input, $X$, to an encoding matrix, $H$, using a deep convolutional neural network followed by a reshaping of the image height dimension into the channels dimension. 
Next, a \textbf{contextual encoder}, $g: \mathcal{H} \mapsto \mathcal{C}$, computes a contextualized representation matrix, $C$, from $H$ using a neural sequence model, parameterized by either an LSTM or Transformer \citep{hochreiter1997long,vaswani2017attention}.
We describe both the design of $f$, which determines the output size of the convolutional encoding space $\mathcal{H}$, and $g$ in Section~\ref{sec:experiment_details}.
Together, both the convolutional and contextual layers form the encoder of text line images  used for downstream document transcription. 
Ideally, $f$ will capture the underlying visual appearance of distinct character classes, while $g$ will discover linguistic correlations between these classes. 

\subsection{Masking}
During pre-training, we replace randomly sampled, non-overlapping segments of $H$ with a learned mask embedding vector prior to computing contextualized representation matrix $C$.
We train the model to distinguish the masked region from a foil using the contrastive loss presented in the next section.

\subsection{Pre-training Objective}\label{sec:pretrain_objective}
We use the following self-supervised contrastive loss whose variants have demonstrated success in self-supervised representation learning \citep{oord2018representation,baevski2020wav2vec}: 
\begin{align*}
\mathcal{L}_U(c_{t}) = -\log \frac{\exp{\big(s(c_{t},h_{t}}) \big)}{\sum_{t'} \exp{\big(s(c_{t}, h_{t'}) \big)}}
\end{align*}
\\ Here, $c_{t}$ (depicted in Figure~\ref{fig:pretrain_model_fig_basic}) is the contextual encoder's output representation of the \textit{masked} line image at position $t$. 
Similarly, $h_{t}$ (also depicted in Figure~\ref{fig:pretrain_model_fig_basic}) is the \textit{convolutional} encoder's output representation of the \textit{masked region itself}. 
Further, $s(c, h)$ represents a scoring function that computes the similarity between representation vectors $c$ and $h$.
We use the cosine similarity similar to \citet{baevski2020wav2vec}, but compute it only raw vectors, instead of the raw vectors and quantized vectors.
The cross-entropy loss requires the model to distinguish the representation of the true masked region, $h_{t}$, from distractor representations: the convolutional encodings of other segments, $h_{t'}$ with $t'\ne t$.

\subsection{Fine-tuning Objective}\label{sec:finetune_objective}

After learning pre-trained representations, we add the randomly initialized, fully connected character vocabulary projection layer to the top of our context encoder network (top right of Fig.~\ref{fig:pretrain_model_fig_basic}) and perform supervised training using the Connectionist Temporal Classification (CTC) objective \citep{graves2006connectionist,graves2012offline,baevski2020wav2vec} with transcribed data.
CTC is a commonly used loss function for supervised training in speech and handwriting recognition systems.
In this case, CTC is used to marginalize over all monotonic alignments between the sequence of input visual representations and the observed ground truth output sequence of characters.

\section{Datasets} \label{sec:datasets}

\begin{figure}
    \centering
    \frame{\includegraphics[trim=0 0 0 0,clip,width=\linewidth]{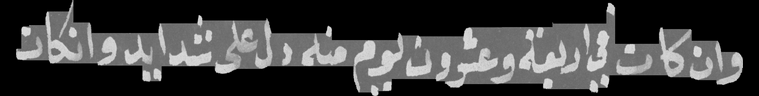}}
    \frame{\includegraphics[trim=0 0 0 0,clip,width=\linewidth]{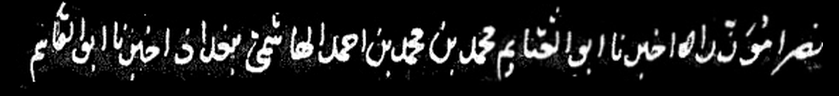}}
    \frame{\includegraphics[trim=0 0 0 0,clip,width=\linewidth]{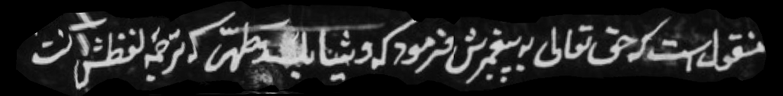}}
    \frame{\includegraphics[trim=0 0 0 0,clip,width=\linewidth]{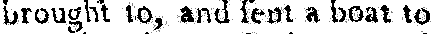}}
    \frame{\includegraphics[trim=0 0 0 0,clip,width=\linewidth]{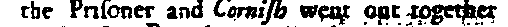}}
    \caption{Assortment of cropped, grayscale line images from a selection of our datasets, as extracted by annotators. From top to bottom, RASM 2019 \citep{keinanschoonbaert_rasm_2019}, Attar-Mubhij,  \d{H}uliyya, Trove \citep{holley2010trove}, Old Bailey \citep{shoemaker2005digital}. The Arabic-script line images are shown pre-binarization, while the English line images come binarized.}
    \label{fig:line_images}
\end{figure}

In this section, we describe both unlabeled pre-training and labeled fine-tuning/testing datasets used in our experiments. 

\subsection{Islamicate Manuscripts}

First, we introduce a variety of Islamicate manuscript datasets selected for both their uniquely different domain content (e.g., scientific to legal to religious) and their visually distinct scribal handwriting style. 

\paragraph{HMML Pre-train}
Through a collaboration with the Hill Museum and Manuscript Library (HMML), we obtain about 100 early modern, mostly Syrian, naskh\footnote{\url{https://en.wikipedia.org/wiki/Naskh_(script)}} manuscripts dating from 1600--1775 with some voweling, but with ornamentally voweled texts excluded (i.e., texts in which every single vowel and orthographic feature is included, usually for ornamental reasons). 
We filter out manuscripts with extensive marginalia, figures, or tables, though some marginal notes and other elements (e.g., seals, interlinears) are still present.
This results in a dataset containing roughly 750,000 unlabeled line images.

\paragraph{HMML Fine-tune} We obtain professional transcriptions for 115 line images from a 4-page held-out subset of the above HMML Pre-train dataset. 
This dataset is designed for in-domain fine-tuning/testing experiments with our pre-trained models.

\paragraph{RASM 2019} For the ICDAR 2019 Competition on Recognition of Historical Arabic Scientific Manuscripts, the British Library released 2,164 manually transcribed line images from scientific manuscripts written in various scribal hands \citep{keinanschoonbaert_rasm_2019}.
RASM 2019 has become a popular benchmark for Arabic-script handwriting recognition due to its relatively large amount of supervised data for the task.

\paragraph{Attar-Mubhij} 
An Arabic-language legal text containing 190 professionally transcribed line images.

\paragraph{\d{H}uliyya} A professionally transcribed, 229-line Persian, nasta'līq\footnote{\url{https://en.wikipedia.org/wiki/Nastaliq}} devotional text by an early modern scholar containing mostly Arabic-language prayers.

\subsection{Early Modern English Printed Works}

Next, we describe several English book and newspaper datasets used in our experiments that were originally printed in early modern England and Australia.

\paragraph{EEBO Pre-train} We harvest 750,000 unlabeled line images from a randomly sampled collection of document images from Early English Books Online (EEBO),\footnote{\url{https://www.proquest.com/eebo}} which contains ``almost every work printed in the British Isles and North America, as well as works in English printed elsewhere from 1470-1700.'' 

\paragraph{Trove} A dataset of historic Australian newspapers (c. 1803--1954) from the National Library of Australia \citep{holley2010trove}.
We use the manually transcribed version totaling 450 lines  \citep{berg2013unsupervised}.

\paragraph{Old Bailey} A manually transcribed set of 20 documents printed 1716--1906, consisting of 30 lines per document, taken from \citet{berg2014improved}.
\citet{shoemaker2005digital} compiled the documents, which describe proceedings of London's Old Bailey Courthouse.

\subsection{Line Extraction}
Since our model processes individual line images of a document, we use \citet{kiessling2020modular}'s line extraction method to automatically segment page images into their component text line images for at-scale collection of the pre-training datasets.
We find and discard poorly extracted line images outside an empirically determined pixel width-to-height ratio range of 6--23.

\section{Results} \label{sec:results}

In this section, we present document transcription results for both Islamicate manuscripts and early modern English works introduced in Section~\ref{sec:datasets}.
We compare performance against supervised and unsupervised prior work, and investigate the impact of pre-training/fine-tuning dataset sizes.

\subsection{Experimental Details} \label{sec:experiment_details}
\paragraph{Encoder} For all experiments, we binarize the line images and scale them to a height of 96 pixels, but allow them to vary in width.
We base our CNN architecture on the Kraken OCR system \citep{kiessling2019kraken}: two rectangular $4\times2$ kernels first process the input image, each followed by a Leaky ReLU activation and Group Norm.
Two max pooling operations are applied, one before and one after the final $3\times3$ convolutional layer kernel, with kernel sizes/strides of $4\times2$/$1\times2$ for both.
The first kernel uses a stride of $4\times2$ and the final two both use $1\times1$.
The convolutional hidden dimensions are 64, 128, and 256.
We use a 3-layer BiLSTM for our contextual encoder with a hidden size of 512.
Models are implemented in PyTorch \citep{paszke2019pytorch} and Fairseq \citep{ott2019fairseq}.

\paragraph{Pre-training} During pre-training, we perform a non-exhaustive grid search over masking probability and length using 75k lines of data.
We determine $p=0.5/p=0.65$ to perform best for Islamicate manuscript/English print with a non-overlapping segment length of 12 time steps.
We ensure that 8 time steps are between each non-overlapping segment.
A maximum of 100 time steps are sampled and used as foils in the denominator of the loss from Sec.~\ref{sec:pretrain_objective}.
We use the same learning rate scheduler and Adam optimizer from \citet{baevski2020wav2vec} that warms up for the first 8\% of updates to a learning rate of 5e-4 and linearly decays it afterwards.

\begin{table}[t]
    \centering
    \begin{adjustbox}{width=\columnwidth,center}
    \begin{tabular}{l c c c c}
    \toprule
        \multicolumn{5}{c}{30 Lines for Supervised Fine-tuning}\\
              \toprule
        & \multicolumn{4}{c}{Fine-tune/Test Dataset CER ($\downarrow$)} \\\cmidrule(lr){2-5}
        \# Lines Pretrain & HMML-F & RASM & Attar-Mubhij & \d{H}uliyya \\
      \midrule
        0    & 51.0 & 68.9 & 60.4 & 70.3 \\
        75k  & 22.7 & 46.1 & 30.4 & 52.9 \\
        750k & \bf 14.8 & \bf 36.2 & \bf 23.7 & \bf 45.5 \\
        \toprule
        \multicolumn{5}{c}{90 Lines for Supervised Fine-tuning}\\
        \toprule
        & \multicolumn{4}{c}{Fine-tune/Test Dataset CER ($\downarrow$)} \\\cmidrule(lr){2-5}
        \# Lines Pretrain & HMML-F & RASM & Attar-Mubhij & \d{H}uliyya \\
        \midrule
        0    & 36.9 & 61.7 & 36.8 & 52.5 \\
        75k  & 15.2 & 34.4 & 20.8 & 37.5 \\
        750k & \bf 10.0 & \bf 25.9 & \bf 15.0 & \bf 28.3\\
    \bottomrule
    \end{tabular}
    \end{adjustbox}
    \caption{30 line and 90 line supervised fine-tuning, tested on held-out portion of fine-tuning dataset. Character error rate (CER) is reported.}
    \label{tab:arabic_finetuning30}
\end{table}

\begin{table}[t]
    \centering
    \begin{tabular}{l c c}
    \toprule
    \multicolumn{3}{c}{Baselines}\\
    \toprule
        & \multicolumn{2}{c}{Test Dataset CER ($\downarrow$)} \\\cmidrule(lr){2-3}
        System & Trove & Old Bailey \\
    \midrule
    Google Tesseract & 37.5 & -  \\
    ABBYY FineReader & 22.9 & - \\
    Ocular & 14.9 & 14.9 \\
    Ocular Beam & 12.9 & 10.9 \\
    Ocular Beam-SV & \bf 11.2 & \bf 10.3\\
    
    \toprule
        \multicolumn{3}{c}{30 Lines for Supervised Fine-tuning}\\
        \toprule
        & \multicolumn{2}{c}{Test Dataset CER ($\downarrow$)} \\\cmidrule(lr){2-3}
        \# Lines Pretrain & Trove & Old Bailey \\
      \midrule
        0    & 66.8 & 56.6  \\
        75k  & 15.1 & 16.8 \\
        750k & \bf 13.9 & \bf 12.9 \\
        \toprule
        \multicolumn{3}{c}{90 Lines for Supervised Fine-tuning}\\
        \toprule
        & \multicolumn{2}{c}{Test Dataset CER ($\downarrow$)} \\\cmidrule(lr){2-3}
        \# Lines Pretrain & Trove & Old Bailey  \\
        \midrule
        0    & 23.7 & 13.1 \\
        75k  & 10.2 & 5.6 \\
        750k & \bf 8.6 &  \bf 5.2 \\
    \bottomrule
    \end{tabular}
    \caption{30 line and 90 line supervised fine-tuning, tested on held-out portion of each fine-tuning dataset. Character error rate (CER) is reported ($\downarrow$). Baselines are duplicated from \citet{berg2014improved}.}
    \label{tab:eebo_finetuning30}
\end{table}

\paragraph{Fine-tuning} During fine-tuning, we use a tri-stage learning rate schedule with the Adam optimizer, which warms up the learning rate to 5e-4 during the first 10\% of updates and decays it linearly by a factor of 0.05 for the final 50\% of training.
We only update the fully connected layer for the first 200 epochs of training and then proceed to update the contextual encoder as well.
These optimization choices are inspired by \citet{baevski2020wav2vec}.
We use a small batch size of 8 and train for a maximum of 700 epochs with the CTC loss (Sec.~\ref{sec:finetune_objective}).
We use greedy decoding after removing the CTC's blank token and do not use any external language model.
For Islamicate manuscript experiments we perform NFD unicode normalization. 

\paragraph{Fine-tune/Test Splits}
For Islamicate manuscript datasets, we hold out 10\% of RASM 2019 for testing and one page each of HMML Fine-tune, Attar-Mubhij, and \d{H}uliyya.
For English print datasets, we use the same test splits as \citep{berg2014improved} for fair comparison and fine-tune on the validation set of each dataset.

\subsection{Islamicate Manuscripts}

In Table~\ref{tab:arabic_finetuning30}, we present supervised fine-tuning results on in-domain subsets of each dataset limited to 30 and 90 lines (roughly 1 and 3 pages of data, respectively).
Each row represents a different set of encoder parameters, which we use to initialize the fine-tuning experiments.
Zero lines represents a randomly initialized encoder, while 75k and 750k settings use the encoder parameters pre-trained with our lacuna reconstruction objective on different orders of magnitude of unlabeled HMML Pre-train line images.

The first thing we can observe is the extremely high character error rates for the randomly initialized models, especially in the 30-line setting.
Access to 2 more pages of data (in the 90-line setting) improves results for this setting in the Arabic-language legal text Attar-Mubhij, but does not seem to help much for RASM 2019, a larger collection of scientific manuscripts.
This is probably due to the higher amount of diversity in content and style in this benchmark dataset for Arabic-language HTR.
Seemingly, without any signal from pre-training and only tens of lines of transcribed data, the model is unable to learn a sufficient visual encoder for the large variety of scribal hands and scripts observed in the manuscripts (examples shown in Fig.~\ref{fig:line_images}).
Pre-training on just 75k lines halves the error rate for Attar-Mubhij in the 30-line setting.
Meanwhile, 750k lines of pre-training reduces the Attar-Mubhij CER from 60.4 to 23.7.

\begin{figure*}[t]
\centering
\begin{subfigure}{.5\textwidth}
  \centering
  \includegraphics[trim=1 1 2 2,clip,width=1\textwidth]{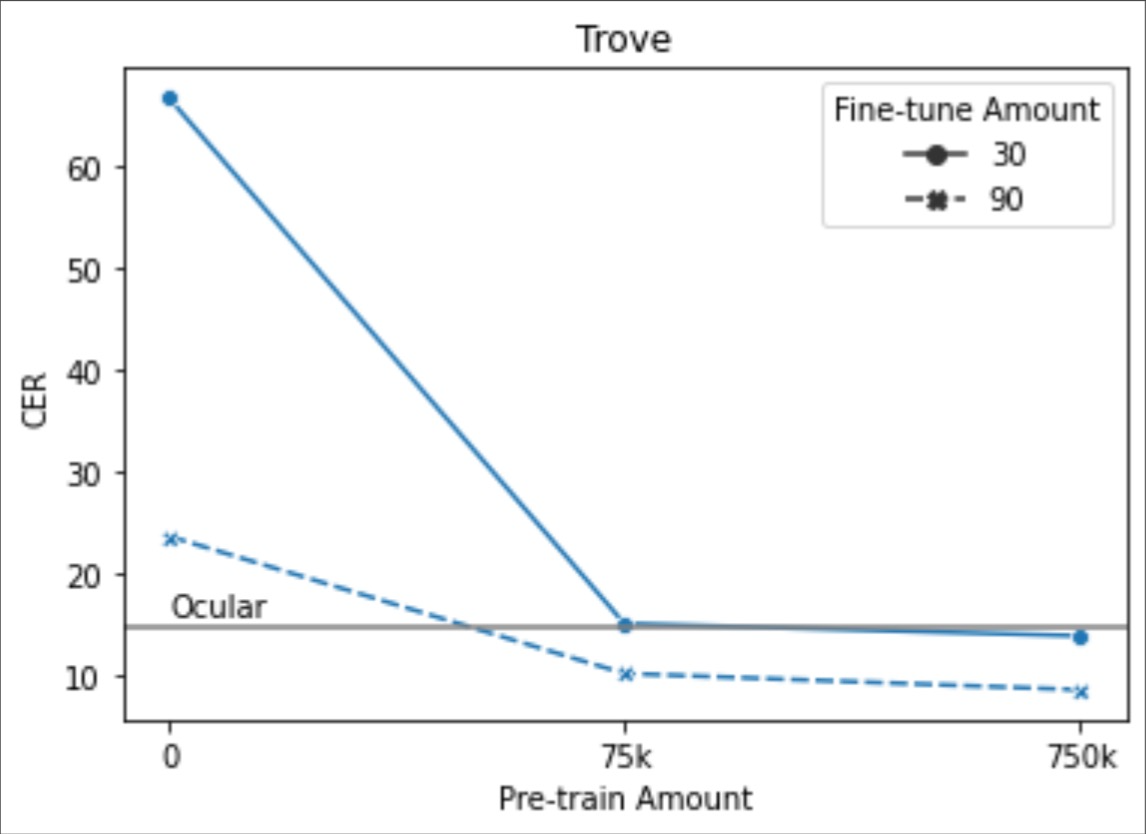}
\end{subfigure}%
\begin{subfigure}{.5\textwidth}
  \centering
  \includegraphics[trim=1 1 2 2,clip,width=1\textwidth]{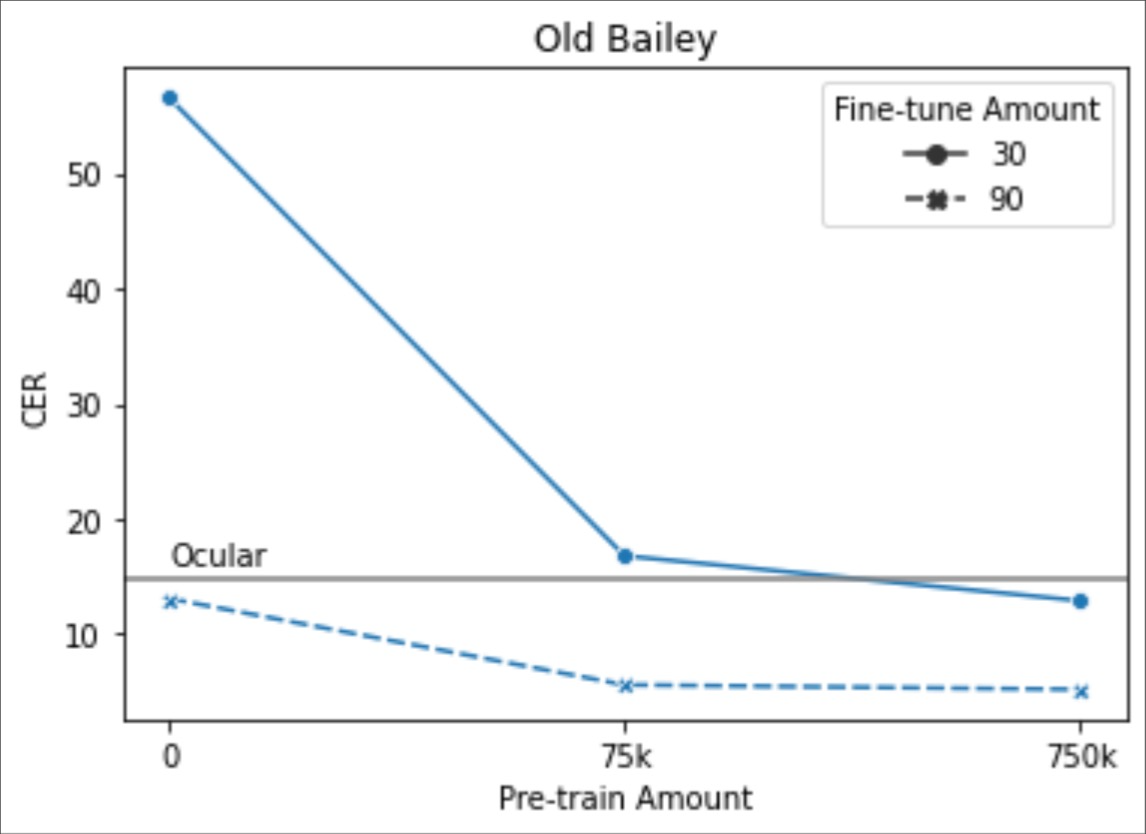}
\end{subfigure}
\caption{Effect of varying pre-train and fine-tune data against the most comparable Ocular baseline without beam search (topmost Ocular setting in Table~\ref{tab:eebo_finetuning30}) for Trove \textbf{(left)} and Old Bailey \textbf{(right)}.}
\label{fig:vary_english}
\end{figure*}

The HMML Fine-tune dataset (HMML-F in Table~\ref{tab:arabic_finetuning30}) has the largest relative error rate difference between the pre-trained models and models without pre-training.
Errors are reduced by about 55\% for 75k-30, 70\% for 750k-30, 58\% for 75k-90, and 73\% for 750k-90, which is at least 10 points higher than other datasets on average.
Since manuscripts in HMML-F are sourced from the same library as the HMML Pre-train dataset, the results suggest that in-domain pre-training data provides an advantage over the other documents from different collections.
Regardless, our approach's improved performance on 30-line settings compared to the supervised 90-line results trained from scratch across all datasets is impressive and shows promising generalization ability.

\subsection{Early Modern English Printed Works}

In Table~\ref{tab:eebo_finetuning30}, we present supervised fine-tuning results on in-domain subsets of each dataset limited to the same 30 and 90 line settings as in the Islamicate manuscript experiments.
Our first observation is that the randomly initialized encoder from the 0-line pre-train setting sees a much larger improvement from 30 to 90 lines of supervised fine-tuning data than the Islamicate manuscript experiments.
We hypothesize that this is a result of the more similar and repeated glyph shapes on printed data compared to handwritten data, which makes learning of the visual encoder easier.
Still, pre-training the visual encoder cuts CER across both datasets, though we do see a slightly bigger relative error rate reduction when fine-tuning on Trove compared to Old Bailey.

In Figure~\ref{fig:vary_english}, we visualize the effect of more pre-train/fine-tune data using the results from  Table~\ref{tab:eebo_finetuning30}.
We note that we use the Ocular baseline without beam search (i.e., not Ocular Beam and Ocular Beam-SV, the best performing baseline models) for this comparison, for a better comparison with our greedy-style decoding, which also does not use a language model.
All pre-training methods are able to improve over Ocular in the 90-line setting, but the visual encoder pre-trained on 750k lines also improves over Ocular with only 30 lines of transcribed data.

\subsection{Conclusion}

In this paper, we proposed a two-phase pre-train/fine-tune approach for document transcription and applied it to historical documents in low-resource settings.
Our pre-training strategy, inspired by reconstructing missing information in documents, or lacuna, uses hundreds of thousands of unlabeled line images to learn rich visual language representations.
After supervised fine-tuning on tens of transcribed line images, we showed large character error rate reduction on both Islamicate manuscripts exhibiting major script and style variation and improved over the unsupervised state-of-the-art OCR system on early modern English printed works.
We estimate that our approach could save human annotators significant amounts of time and enable more distant readings of library collections.

\bibliography{anthology,custom}
\bibliographystyle{acl_natbib}

\end{document}